\def\year{2018}\relax
\documentclass[letterpaper]{article} 
\usepackage{aaai18}  
\usepackage{times}  
\usepackage{helvet}  
\usepackage{courier}  
\usepackage{url}  
\usepackage{graphicx}  
\usepackage{mathrsfs}
\usepackage{multirow}
\usepackage{pgfplots}
\pgfplotsset{compat=1.14}

\begin{document}
%
\title{Automatic Generation of Language-Independent Features\\ for Cross-Lingual Classification}
\author{Sarai Duek \and Shaul Markovitch\\
\{sarai, shaulm\}@cs.technion.ac.il\\
Technion -- Israel Institute of Technology\\
32000 Haifa, Israel\\
}


\maketitle
\begin{abstract}
Many applications require categorization of text documents using predefined categories. The main approach to performing text categorization is learning from labeled examples. For many tasks, it may be difficult to find examples in one language but easy in others. The problem of learning from examples in one or more languages and classifying (categorizing) in another is called cross-lingual learning. In this work, we present a novel approach that solves the general cross-lingual text categorization problem. Our method generates, for each training document, a set of language-independent features. Using these features for training yields a language-independent classifier. At the classification stage, we generate language-independent features for the unlabeled document, and apply the classifier on the new representation.

To build the feature generator, we utilize a hierarchical language-independent ontology, where each concept has a set of support documents for each language involved.
In the preprocessing stage, we use the support documents to build a set of language-independent feature generators, one for each language. The collection of these generators is used to map any document into the language-independent feature space.

Our methodology works on the most general cross-lingual text categorization problems, being able to learn from any mix of languages and classify documents in any other language. We also present a method for exploiting the hierarchical structure of the ontology to create virtual supporting documents for languages that do not have them.  
We tested our method, using Wikipedia as our ontology, on the most commonly used test collections in cross-lingual text categorization, and found that it outperforms existing methods.
\end{abstract}

\section{Introduction}
\label{chapter:Introduction}

Many applications require categorization of text documents using predefined categories. The main approach to performing text categorization is learning from labeled examples. Typically, our task is categorizing documents in a specific language, when the labeled examples are in the same language. We call such setups Single-Language 
Text Categorization (SLTC).

In today's global world, the labeled documents and the documents to be categorized are written in multiple languages. 
Such a setup is called Cross-Lingual Text Categorization (CLTC). Solving CLTC problems is hard because mostly, the features used for learning and classifying are based on the words in these documents, and are, therefore, language-dependent; classifiers trained on one language cannot be applied to documents in another. Also, we cannot mix training documents written in multiple languages.

Most existing approaches for solving CLTC problems convert all the documents into one language, thus turning the CLTC task into an SLTC task. The conversion is performed either by using a black-box translator such as Google Translate (for example, \cite{Wan:2009:CCS:1687878.1687913}), or by building conversion tools based on bilingual dictionaries \cite{steinberger2006exploiting}, bilingual thesauri \cite{Ni:2011:CLT:1935826.1935887} or parallel corpora \cite{mogadala2016bilingual,xu2017cross}.

There are several weaknesses to the translation-based approaches.
First, translation is considered to be a much more difficult task than text categorization. Thus, using such methods means reducing a problem to one that is much harder.
Achieving high-quality translation requires deep semantic understanding of the documents, which is still beyond the state of the art. It is true that some progress has been made recently in machine translation, by using deep neural networks, but the problem of translation is still far solved. Second, some works use Google Translate as their translator. This is problematic given that it is a proprietary tool. Google Translate is used as a black box, without understanding of its inner mechanism. Also, translation is usually a rather computationally-intensive process. 

In recent years, several works tackled CLTC problems using a different approach. An interim feature space is defined, and documents from all languages are mapped into this feature space (for example, \cite{Ni:2011:CLT:1935826.1935887,Blei:2003:LDA:944919.944937}). Most of these interim spaces are based on semantic concepts derived from various available semantic resources \cite{sogaard2015inverted,franco2014knowledge,Gliozzo:2006:ECC:1220175.1220245}.

In this work, we present a general framework for solving CLTC problems by generating language-independent features, thus projecting documents into an interim feature space. Using such features allows us to work with any mix of languages in our training and testing sets. We assume a given (potentially hierarchical) ontology consisting of semantic concepts. These concepts will be used as the language-independent features.

We also assume that each concept has an associated descriptive text for each language involved in the CLTC task. Our algorithm uses these texts to build a single-language interpreter for each language, which enables mapping of documents in the various languages into these semantic features. 

During training, the set of examples is transformed into feature vectors using the generated language-independent features. We can then apply any learning algorithm to produce a classifier. Note that the classifier is language-independent as it uses only language-independent features. When a document is to be classified, we first map it into the language-independent feature space, and then apply the classifier on it.
If we are given a hierarchy over these concepts, we can use its links to generate more abstract features. 

In many cases, some concepts in the given ontology do not have supporting documents in some of the languages in the CLTC task.
We introduce a novel algorithm that uses the hierarchy to generate a language-specific virtual support document 
for a concept, when the concept does not have such a support document for that language. 

We implemented the methodology described above and evaluated it on various CLTC setups. We show the generality and flexibility of our framework by applying it to CLTC setups where documents of multiple languages are presented in both the training and testing sets. We also compare the performance of our algorithm to various existing baselines and show its superior performance.

\section{Problem Definition}
\label{Section:Problem Definition and UCLTC}

In this section we define the problem of Cross-Lingual Text Classification (CLTC). First, let us define the general Text Classification (TC) problem. Let $D$ be a set of documents in natural language text. Let $K= \{ k_1,...,k_r \} $ be a finite set of categories, and let us assume there exists an unknown function $\varphi ^*:D \rightarrow K$ that classifies each document in D as belonging to exactly one category in K. 

Assume we are given a subset of documents $D^e \subseteq D$ for which the function $\varphi ^*$ is known, and a subset of documents  $D^c \subseteq D\setminus{D^e}$ for which $\varphi ^*$ is unknown. Our goal is to estimate $\varphi ^*$ for any document in $D^c$.

An inductive algorithm $\cal{A}$ receives a training set $ E=\{ \langle d, \varphi ^* (d)\rangle : \forall d \in D^e\}$, and induces a classifier function $\varphi:D \rightarrow K$ that estimates $\varphi ^* $. We can then use $\varphi $ to classify any given document in $D^c$.  

In traditional TC, the documents in $D^e$ and $D^c$ are all in the same language. In CLTC problems, however, the language of the documents in each of the two sets can vary, leading to several interesting CLTC setups. Let $\mathscr{L}=\{ l_1,\ldots ,l_t \}$ be a set of natural languages. We define an operator $l:D \rightarrow \mathscr{L}$ that identifies the language of each given document. We can now define an operator $L$ that, given any set of documents $D'$, returns the set of their languages:

\[ L(D')=\{ l(d): \; \forall \, d  \in D' \} \subseteq \mathscr{L} \] 

Given a CLTC task $\left\langle D^e,D^c \right\rangle$, we denote the set of languages of the task as $L_{D^e, D^c}=\{ L(D^e)\cup L(D^c)\}$. We can now list some of the CLTC setups:
\begin{enumerate}
   \item \textbf{CLTC1:} The traditional TC setup, where all documents from $D^e$ and $D^c$ are in the same language:
$ |L(D^e)|=1,\quad |L(D^c)|=1\quad and\quad L(D^e)=L(D^c) $
   \item \textbf{CLTC2:} The most common  CLTC setup, where all the documents in $D^e$ are in one language, and all of the documents in $D^c$  are in another language:
$ |L(D^e)|=1,\quad |L(D^c)|=1\quad and\quad L(D^e)\neq L(D^c) $
   \item \textbf{CLTC3:} $D^e$ may be multilingual while the documents in $D^c$ are all in a single language: 

$ |L(D^e)|>1,\quad |L(D^c)|=1\quad \mbox{and potentially } L(D^c) \subseteq L(D^e) $

   \item \textbf{UCLTC:} The most general setup , where $D^e$ and $D^c$ are both multilingual, and can either intersect or not. We call this case the Unrestricted Cross-Lingual Text Categorization (UCLTC) setup:

$ |L(D^e)|\geq 1,\quad |L(D^c)|\geq 1\quad, \mbox{and potentially }  L(D^e)\cap L(D^c) \neq \emptyset $

\end{enumerate}

In this work we present a solution to the most general CLTC problem -- UCLTC. Our algorithm can accept any mix of languages in the training set and can classify documents in any language.

\section{Generating Language-Independent Features}
\label{Section:General Framework for Solving UCLTC}
In this section, we present a general framework for solving unrestricted cross-language text classification problems. First, we present a general architecture for solving UCLTC tasks. Then we describe each component of our architecture and its associated algorithms. Finally, we show how we employ Wikipedia to instantiate our algorithm.

\subsection{Solution Components}
\label{Section:Solution Components}

Our methodology is based on transforming language-dependent text to language-independent features in a conceptual space. In this space, we are able to express conceptual ideas in a rich and deep form of representation, which is also suitable for learning algorithms. 

Our solution comprises two principle components:
\begin{enumerate}
    \item An ontology: A large set of language-independent concepts $ C= \{ c_1,\dots ,c_k \} $ that is wide and general enough to properly describe documents in $D$. Each concept will serve as a Language-Independent Feature (LIF). 
    \item For each language $l \in L_{D^e, D^c}$, a feature generator $LIFG^l: D \rightarrow 2^C$ that generates, for each document $d$ in language $l$, a set of language-independent features $C' \subseteq C$.
\end{enumerate}

\subsection{Solution Overview}
\label{Section:The CLTC Solution Process}
We define a language-independent feature generator $LIFG: D \rightarrow 2^C$ that generates
for each document a set of language-independent features. It selects the correct language-specific generator and applies it to the document: $LIFG(d)=LIFG^{l(d)}(d)$.
We denote the set of all features generated for the set of training documents by $\mathcal{G}$:
\[\mathcal{G} = \bigcup_{d \in D^e}  LIFG^{l(d)}(d)\]

During training, each example is mapped into the feature space defined by $\mathcal{G}$, 
yielding a training set represented by language-independent features. We then apply a learning algorithm on this training set to induce a language-independent classifier. 

When a test document $d$ is given to the classifier, it first maps it into the feature space defined by  $\mathcal{G}$, using $LIFG^{l(d)}$, and then classifies it using these language-independent features.



\subsection{Building a Language-Independent Feature Generator}
\label{Section:Building Language-Independent Feature Generator}

Our solution requires an ontology and, for each language, a feature generator that maps documents in that language into the ontology. In the following subsections we describe our methodology for building these generators.

\subsubsection{Multilingual Knowledge Resource}
\label{Section:Requirements on a Multilingual Knowledge Resource}

To build a feature generator $LIFG^l$ for each language $l \in L_{D^e, D^c}$, we need a multilingual knowledge resource that satisfies the following requirements:

\begin{enumerate} 
	\item Has a set of basic concepts $C_{Basic}$ that will serve as our ontology. 
	\item For each concept $c \in C_{Basic}$, for each language $l \in  L_{D^e, D^c}$, has a support set of documents. We denote the set of support documents for concept $c$ in language $l$ by $s(c,l)$. 
  \item Has, potentially, a set of meta-concepts, $C_{Meta}$, and an is-a hierarchy $H$ over $C_{Meta} \cup C_{Basic}$:
  \[H=\{ (c, c'): \; c\in C_{Meta}, c'\in C_{Meta}\cup C_{Basic} \}\]
$H$ is assumed to be a Directed Acyclic Graph (DAG).
We now extend the definition of supporting documents for $c \in  C_{Meta}\cup C_{Basic}$ and language $l$ to be the multiset of all supporting documents in language $l$ of $c$ and all of its descendants:
  
  \[
    S(c, l)=\left\{
                \begin{array}{ll}
                  \biguplus_{c'\in Children(c)} S(c',l) &  c\in C_{Meta}\\
                  s(c, l) &  c\in C_{Basic}
                \end{array}
              \right.
  \]
Note that we use a multiset of documents rather than a set, which increases the weight of a document $d$. Document $d$ supports a concept $c'$, which has multiple paths to $c$ (indicating that $c'$ has multiple aspects in our ontology).
\end{enumerate}

Although we can perform feature generation with a flat set of concepts, having a hierarchical ontology adds a powerful generality perspective to the representation of text. 

\subsubsection{Constructing the Feature Generator}
\label{Section:Building a Language-Independent Feature Generator for a Specific Language}

Our method for generating language-independent features for a document in a specific language $l$ is based on Explicit Semantic Analysis (ESA) \cite{Gabrilovich:2009:WBS}. ESA builds a single-language semantic interpreter $SI^l$ that maps text in $l$ to a vector of weighted concepts in $C_{Basic}$.  For each $c \in C$, it first analyzes its set of supporting documents $S(c,l)$ to compute the TF.IDF value of each word appearing there. The analysis yields a matrix where each row stands for a concept, and each column stands for a word. Inverting this matrix yields the desired semantic representation of words: Each row is a vector of weights indicating the strength of association between the word and the concepts.
Given a document $d$ in language $l$, $SI^l$ returns a vector that is the centroid of the vectors of the individual words within it.

Given $SI^l$, we can now define our single-language feature generator $LIFG^l$. 
Let $SI^l(d)=\{ \left\langle c_i, w_i \right\rangle\}$ be the semantic interpretation of document $d$. $LIFG^l(d)$ returns the $k$ concepts $c_i$ with the highest weights $w_i$. To evaluate a generated feature $c_i$ on a given document, we first considered using its TF.IDF value, $w_i$. We found, however, that due to the different number of the terms each semantic interpreter holds for each language and the language characteristics, the TF.IDF ranges in the semantic interpretation vectors differ from language to language, causing range differences in the final weights. These differences distort the learning process and reduce performance. We, therefore, decided to use binary values, which improved performance. An added benefit of using binary representation is the reduced memory usage of binary vectors and their reduced computational cost. 

Let $C^k \subseteq C_{Basic}$ be the $k$ basic features generated by $LIFG^l$ for document $d$. If our ontology is hierarchical, our feature generator enriches this set with more abstract concepts in $C_{Meta}$. Specifically, for each generated $c_i \in C^k$, it generates all its ancestors to $m$ (set to 3 in our experiments) levels above $c_i$. Using the hierarchy allows the learning algorithm to generalize, using multiple levels of abstraction.

As our feature generator can potentially produce a very large set of conceptual features, we employ traditional feature selection based on information gain to make the generated set of features more manageable. We also filter out meta-features that are not ancestors of at least two different $C_{Basic}$ features in $LIFG^l(d)$. 


\subsection{Constructing Virtual Support Documents}
\label{Section:Constructing}
Our feature generation methodology requires a nonempty set of supporting documents $S(c,l) \neq \emptyset$ for each $c \in C_{Basic}$, for each $l \in L_{D^e,D^c}$. 
Constructing such a complete multilingual knowledge resource is, however, time consuming and requires native knowledge in all languages in $L_{D^e,D^c}$. It is also very difficult to maintain such a complete multilingual resource. It is, therefore, inevitable that the multilingual knowledge resource will lack supporting documents in some languages. For example, when using Wikipedia as a knowledge resource, we saw that there are (as of September 2017) about 5.5M articles in English, but only about 2 million in German and French, and 1.3 million in Spanish. Therefore, many concepts are likely to have support articles in English but not in other languages.

If $S(c, l)=\emptyset$ for some concept $c$ and language $l \in L(D^e)$, the feature generator will not be able to generate $c$ for the training documents of language $l$, and will not be able to use $c$ for generalizations. 
If it is empty for $l \in L(D^c)$, a classifier that uses $c$ will encounter a missing value. To avoid these problems, we devised a method for automatic construction of missing supporting documents. We refer to these constructed documents as virtual support documents.

Our method for construction of virtual supporting documents is based on the hierarchical structure of the ontology. Usually, each $c \in C_{Basic}$ has multiple direct parents.
We base our algorithm on the observation that multiple parents reflect multiple aspects of $c$, therefore, inheriting the dominant terms in the supporting documents of $Parents(c)$ is likely to be a good textual representation of the concept.

Let $c \in C_{Basic}$ be a concept with an empty set of support documents in language $l$.
We define $Ancestors(c,i)$ to be the set of all ancestors of $c$ with distance of at most $i$ edges from $c$. Thus, $Ancestors(c,1)$ are all the direct parents of $c$, and $Ancestors(c,2)$ are all the grandparents of $c$ and so on. We go up the hierarchy until there is a sufficient number of support documents:
\[
j = \min \{\,i \,| \sum_{c' \in Ancestors(c,i)}S(c',l) \geq p\}
\]
where $p$ is a system parameter.

We now select from each ancestor $c' \in Ancestors(c,j)$ the $t$ most prominent terms in the concatenation of $S(c',l)$, and use them to construct a count table representing the new virtual support document for $c$. The terms in this table are the prominent terms in all the ancestors. 

This algorithm for constructing virtual support documents allows us to work with incomplete  multilingual knowledge resources efficiently. 

\section{Empirical Evaluation}
\label{Section:Empirical Evaluation}
We implemented our new framework using Wikipedia as an external knowledge resource and evaluated its performance on various multilingual tasks. 

\subsection{Implementation Details}
\label{Subsection:Implementation Details}
As the datasets used for the experiments include documents in four languages  --
English (E), German (G), French (F) and Spanish (S) (we did not use the Japanese and Italian documents) -- we obtained the Wikipedia dumps form November 2015 for these languages. We retained only pure conceptual articles, filtering out extraneous articles such as disambiguation pages, redirection pages, and pages that catalog concepts such as lists, dates and years. We also removed articles that are too short,
and articles of low importance, indicated by a small number of incoming and outgoing Wikipedia inner links.
After the pruning, about 700K articles were left in the English collection, and about 500K articles in each of the other collections. 

Given a UCLTC task $\langle D^e,D^c \rangle$, we retain every concept that has an article in each of the languages in $L_{D^e,D^c} \subseteq \{ E, G, F, S\}$, allowing mapping from each document in the training and testing sets to the same set of conceptual (language-independent) features. We then build a language-independent feature generator for each languages in $L_{D^e,D^c}$. The system parameter $k$, which specifies the number of concepts representing a term, was set to 5K. This concludes the preprocessing stage.

During training, we first use the $LIFG^l$ constructed in the preprocessing stage to 
generate language-independent features for each training document. 
In addition to these features, we generate meta-features. One potential complication is inconsistency in the hierarchical structure between languages. To overcome such inconsistencies, we assign an edge between two concepts if such an edge exists in at least one language.


At this stage, we apply an inductive learning algorithm on the new representation
and induce a language-independent classifier.When testing, we map the test documents into the same space which, therefore, enables us to use the learned classifier to classify them.

\subsection{Test Collections}
\label{Section:Test collections}

Multilingual datasets are hard to obtain. Here we use the two main collections used in previous works:

\begin{enumerate}
    \item \textbf{Webis-CLS-10 \cite{prettenhofer2011cross}}

A collection of product reviews from the international Amazon shopping site. It contains documents in English, French, German and Japanese, where each document is assigned one of three categories -- music, books and DVD. 

For each of the four languages, the dataset includes 6K training documents, and 6K testing documents. The distribution of categories in the training and testing sets is uniform.
In addition to the training and testing sets, there is unlabeled data in each language that we did not use. There are also translations of all non-English documents into English that we did not use. Recall that the Japanese documents were not included in our experiments.
 
    \item \textbf{Reuters RCV1/RCV2 Multilingual Text Categorization \cite{amaniAUG09}}

A multilingual categorization dataset based on the Reuters news collection. It is composed of news documents in English, French, German, Spanish and Italian divided into six categories. The number of documents in each category varies between languages. There are up to 5K samples from each category in each language. The collection also includes multiview translations of each document in all five languages, which we did not use in this paper. Recall that the Italian documents were not included in our experiments.
\end{enumerate}

\subsection{Comparative Performance of Our Algorithm}
\label{Subsection:Baselines}
We compared our algorithm to six baselines. The baselines we chose use either one or both of the test collections described above. We compared our results with the results of the baselines reported in the cited papers. As there was a variation in the experimental setup used for the various baselines, we replicated these setups, including the induction algorithm used, for our own algorithm when comparing to each baseline. 
Note that, unlike some of the baselines, we used neither unlabeled data nor labeled target data for our experiment. 

We compared our method with the following baselines: SHFR-ECOC, by \citeauthor{zhou2014heterogeneous} (\citeyear{zhou2014heterogeneous}), \textsc{Inverted}, by \citeauthor{sogaard2015inverted} (\citeyear{sogaard2015inverted}), DCI, by \citeauthor{fernandez2016distributional} (\citeyear{fernandez2016distributional}), SHFA, by \citeauthor{li2014learning} (\citeyear{li2014learning}), DMMC, by \citeauthor{zhou2016transfer} (\citeyear{zhou2016transfer}), and BRAVE, by \citeauthor{mogadala2016bilingual} (\citeyear{mogadala2016bilingual}).

Table~\ref{table:amazon} shows the performance of our algorithm on the 
Webis-CLS-10 dataset compared to the baselines that used it, and Table~\ref{table:reuters} shows the performance of our algorithm on the Reuters RCV1/RCV2 Multilingual dataset compared to the baseline algorithms that used it. As the baselines did, we reported the accuracy achieved, except when we compare with \textsc{Inverted}, which reported F1, and so did we.

\begin{table}
\centering
\small
\begin{tabular}{|l|c|c|c|c|c|}
\hline
 Baseline& Source&          Target &        Baseline Results &       LIFG    \\
\hline
\multirow{2}{*}{SHFR-ECOC} 
& E &  F &  $62.09$ & $90.00$ \\
& E &  G &  $65.22$ &  $91.29$ \\
\hline
\textsc{Inverted} & E &  G &  $49.00$ &  $91.00$ \\
\hline
\multirow{2}{*}{DCI} 
& E &  F &  $83.80$ & $90.38$ \\
& E &  G &  $83.80$ &  $92.07$ \\
\hline
\end{tabular}
\caption{CLTC Results on the Webis-CLS-10C Dataset}
\label{table:amazon}
\end{table}

\begin{table}
\centering
\small
\begin{tabular}{|l|c|c|c|c|c|}
\hline
 Baseline& Source&          Target &        Baseline Results &       LIFG    \\
\hline
\multirow{2}{*}{SHFR-ECOC} 
& E &  S &  $72.79$ &  $85.70$ \\
& F &  S &  $73.82$ &  $85.95$ \\
& G &  S &  $74.15$ &  $87.16$ \\
\hline
\textsc{Inverted} & E &  G &  $55.00$ &  $89.00$ \\
\hline
\multirow{2}{*}{SHFA} 
& E &  S &  $76.40$ &  $85.70$ \\
& F &  S &  $76.80$ &  $85.95$ \\
& G &  S &  $77.10$ &  $87.16$ \\
\hline
\multirow{3}{*}{DMMC} 
& E &  F &  $65.52$ &  $88.63$ \\
& E &  G &  $58.23$ &  $89.44$ \\
& E &  S &  $62.64$ &  $85.70$ \\
\hline
\multirow{3}{*}{BRAVE} 
& E &  F &  $82.50$ &  $89.39$ \\
& E &  G &  $89.70$ &  $90.76$ \\
& E &  S &  $60.20$ &  $86.78$ \\
& F &  E &  $79.50$ &  $89.09$ \\
& G &  E &  $80.10$ &  $89.25$ \\
& S &  E &  $70.40$ &  $86.61$ \\
\hline
\end{tabular} 
\caption{CLTC Results on Reuters RCV1/RCV2 Dataset}
\label{table:reuters}
\end{table}

The table show a clear advantage of our algorithm over the others. Note that we could not compute the statistical significance of this advantage, since the raw results of the baselines were not available to us.


\subsection{The Effect of Hierarchical Feature Generation}
\label{Section:The Effect of Hierarchical Feature Generation}
We now test the utility of the hierarchical feature generation, using SVM, on the Amazon Webis-CLS-10 test set. For each learning task, we ran 10 experiments. For each experiment, we randomly chose 150 training samples (50 from each category) for learning, and 6K documents for testing.
Note that we use much smaller training sets in order to leave more space for improvement. 

In Table~\ref{table:Hierarchical} we show a comparison between our base approach without the hierarchical feature generation enhancement and with it. As can be seen, the result improvement is significant -- about 10\% on average. Clearly, abstract features significantly contributes to performance, and should, therefore, be used when available.

\begin{table}
\centering
\small
\begin{tabular}{|l|c|c|c|c|}
\hline
Source&         Target &        LIFG -- w/o $C_{Meta}$&         LIFG -- w/ $C_{Meta}$\\
\hline
E &  F &  $52.63$ &  $62.03$\\
E &  G &  $55.19$ & $63.34$\\
F &  E &  $50.87$ &  $60.49$\\
F &  G &  $49.32$ &  $60.88$\\
G &  E &  $51.06$ &  $59.61$\\
G &  F &  $50.01$ &  $61.84$\\
\hline
\end{tabular}
\caption{The Effect of Hierarchical Feature Generation}
\label{table:Hierarchical}
\end{table}

\subsection{Handling General UCLTC Setups}
\label{Section:Results for UCLTC}
Here we demonstrate our framework's capabilities in solving the four CLTC setups we presented. We experimented on Amazon Webis-CLS-10 using SVM. For each of the three languages, we randomly chose 150 training samples (50 from each category). We used all the 6K testing documents for testing.
Here again, we use much smaller training sets so that we can analyze the differences in performance when using different UCLTC setups.

We tested all combinations of source and target languages for all the CLTC setups.
The results are shown in Table~\ref{tab:CLTC123}.

Let us look, for example, at all the learning tasks where the target language is French.
When training on French documents, the accuracy achieved is $65.7\%$. When training on documents in English, the accuracy is lower ($62.0\%$) but not by much. Training on German documents yields a bit lower accuracy ($59.6\%$). If, however, we enhance the French training documents with German training documents, the accuracy increases to $71.4\%$, which is better than French alone. Enhancing it with English documents yields an even better performance of $73.5\%$.
Using the training documents of the three languages, yields the best performance -- an accuracy of $77.9\%$. Using each of the other two languages as a target language results in similar patterns. 

For the general setup, UCLTC, the results are shown at the bottom of Table~\ref{tab:CLTC123}.
We can see similar patterns to those shown above. With every source language added to the training set, the performance of the testing set (now in two or three target languages) improves.
\begin{table}[h!]
\centering
\small
\begin{tabular}{|l|c|c|c|c|c|}
\hline
Setup& Source& Target& Examples& LIFG (\%)\\
\hline 
\multirow{3}{*}{CLTC1} 
& F & F & 150 & 65.70\\
& E & E & 150 & 67.60\\
& G & G & 150 & 67.10\\
\hline
\multirow{6}{*}{CLTC2} 
& E & F & 150 & 62.00\\
& G & F & 150 & 59.60\\
& F & E & 150 & 60.50\\
& G & E & 150 & 61.80\\
& F & G & 150& 60.90\\
& E & G & 150& 63.30\\
\hline 
\multirow{11}{*}{CLTC3} 
& F, E & F & 300 & 73.50\\
& F, G & F & 300 & 71.40\\
& E, G & F & 300 & 69.60\\
& F, E, G & F & 450 & 77.30\\
& F, E & E & 300 & 74.60\\
& F, G & E & 300 & 67.40\\
& E, G & E & 300 & 75.20\\
& F, E, G & E & 450& 78.40\\
& F, E & G & 300 & 66.80\\
& F, G & G & 300 & 73.00\\
& E, G & G & 300 & 74.80\\
& F, E, G & G & 450 & 77.90\\
\hline
\multirow{28}{*}{UCLTC} 
& F & F, E & 150 & 63.10\\
& E & F, E & 150 & 64.80\\
& G & F, E & 150 & 60.70\\
& F, E & F, E & 300 & 74.00\\
& F, G&	F, E & 300 & 69.40\\
& E, G&	F, E & 300 & 72.40\\
& F, E, G& F, E & 450 & 77.60\\
& F & F, G & 150 & 63.30\\
& E & F, G & 150 & 62.70\\
& G & F, G & 150 & 63.40\\
& F, G & F, G & 300 & 70.20\\
& F, G&	F, G & 300 & 72.20\\
& E, G&	F, G & 300 & 72.20\\
& F, E, G& F, G & 450 & 77.60\\
& F & E, G & 150 & 60.70\\
& E & E, G & 150 & 65.50\\
& G & E, G & 150 & 64.50\\
& F, E & E, G & 300 & 70.70\\
& F, G&	E, G & 300 & 71.10\\
& E, G&	E, G & 300 & 75.00\\
& F, E, G& E, G & 450 & 78.20\\
& F & F, E, G & 150 & 62.40\\
& E & F, E, G & 150 & 64.30\\
& G & F, E, G & 150 & 62.80\\
& F, E & F, E, G & 300 & 71.60\\
& F, G&	F, E, G & 300 & 70.60\\
& E, G&	F, E, G & 300 & 73.20\\
& F, E, G& F, E, G & 450 & 77.90\\
\hline
\end{tabular}
\caption{Results for CLTC1, CLTC2, CLTC3 and UCLTC}
\label{tab:CLTC123}
\end{table}

\subsection{The Utility of Virtual Support Documents}
\label{Section:The Utility of Constructed Documents}

We test the utility of the constructed virtual documents by simulating a situation of missing documents. The protocol of the experiment is:

\begin{enumerate}
\item We sort our set of concepts in decreasing order, by the length of their associated English supporting articles, assuming that the length indicates the importance of the concept.
\item We start with the set of the first 20K concepts in the sorted list.
We perform a learning experiment and measure the accuracy of the classifier.
\item We add the next 1K concepts in the sorted list. For English, we use the original associated articles. For French, we construct articles for these 1K concepts. We then perform a learning experiment again.
\item We repeat this procedure 10 times.
\item For control, we repeat the experiment, but this time use the original French documents.
\end{enumerate}

For these experiments we used the Webis-CLS-10 dataset, learning with 6K English training sampled using SVM, and testing on 6K French testing documents. 

\begin{figure}
\centering
\begin{tikzpicture}
\begin{axis}[
    xlabel={Number of Concepts},
    ylabel={Accuracy [\%]},
    xmin=20000, xmax=30000,
    ymin=89.4, ymax=91,
    xtick={20000, 21000, 22000, 23000, 24000, 25000, 26000, 27000, 28000, 29000, 30000},
    ytick={89.4, 89.6, 89.8, 90.0, 90.2, 90.4, 90.6, 90.8, 91.0},
    legend pos=north west,
    ymajorgrids=true,
    grid style=dashed,
    label style={font=\small},
    tick label style={font=\scriptsize},
]
\addplot[
    color=blue,
    mark=square,
    ]
    coordinates {
    (20000,89.61125)(21000,89.8558)(22000,89.9289)(23000,89.98689)(24000,90.05965)(25000,90.1256879)(26000,90.197586)(27000,90.24569)(28000,90.33987)(29000,90.34258)(30000,90.341125)
    };

\addplot[
    color=black,
    mark=triangle,
    ]
    coordinates {
    (20000,89.61125)(21000,89.89998)(22000,90.02886)(23000,90.07636)(24000,90.12686)(25000,90.19214)(26000,90.24187)(27000,90.40012)(28000,90.51008)(29000,90.62223)(30000,90.71666)
    };
    \legend{Virtual Support Documents, Original Support Documents}
     
\end{axis}
\end{tikzpicture} 

\caption{The Effect of Virtual Document Reconstruction} 
\label{fig:const}
\end{figure}
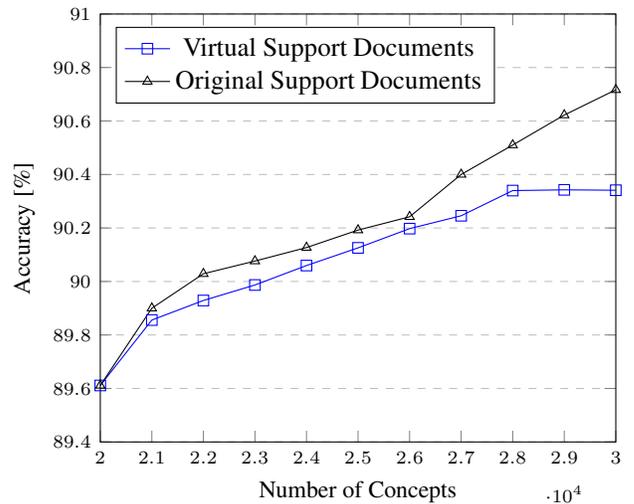

The results are presented in Figure \ref{fig:const}. The plot with the square points shows the performance when using the constructed supporting articles. 
We can see that the constructed articles have positive utility, as the accuracy increases when adding them. For comparison, we show the performance when using the original French documents. 
We can see that the constructed virtual articles are of high quality as their accuracy is very close to that achieved when using the original articles. 
When the proportion of the virtual articles in the target language becomes large, the advantage of the control setup increases. 

\section{Related Work}
\label{Section:Related Work}

Most existing works on cross-lingual text classification assume the CLTC2 setup, where the training documents are written in a \emph{source} language and the testing documents are written in a different \emph{target} language.
Learning in one language and classifying in another requires some method of transformation between these languages. 
Without loss of generality, we will describe all the works as if they receive training documents in one language and testing documents in another language. Some methods require, in addition to the source-language training documents, a (usually small) set of labeled target-language documents. 

We will review related work, grouped by the method they use to overcome the gap between the different languages of the CLTC task: (a) representing all documents (training and testing) in the same feature space, or (b) transforming a model learned using the source language to one that can classify documents in the target language. 


The most obvious approach for solving the CLTC problem is translating the documents so that all the training and testing documents are in the same language, then using common single-language text categorization tools. 

Using this approach, however, transforms the problem into the very difficult task of machine translation. The translation is performed either on the entire documents \cite{Wan:2009:CCS:1687878.1687913,Ling08canchinese}, or on the extracted BOW features \cite{montalvo2007multilingual,shi2010cross}, during the learning or classification phase. When working with the BOW representation, simple translation tools such as bilingual dictionaries \cite{Bel03cross-lingualtext,xu2016cross}, thesauri and nomenclatures \cite{steinberger2006exploiting} can be quite successful. 

One problem in using simple bilingual dictionaries is handling polysemy and synonymy. This problem can be partially solved by building a context-aware bilingual dictionary from parallel corpora. Such parallel corpora, where texts in different languages aligned on a paragraph, sentence or word level, can be used for learning a probabilistic context-based dictionaries \cite{mogadala2016bilingual}. \citeauthor{olsson2005cross} (\citeyear{olsson2005cross}), used such a dictionary while working with English and Czech documents, and experimented by once translating the English documents into Czech and once translating the Czech documents into English, then performing monolingual text categorization.

Some work circumvents the main problem of CLTC by using a black-box machine translation tool, such as Google Translate and others, to translate whole documents \cite{Ling08canchinese,amini2010co}. 
The translated labeled documents are then used, perhaps with additional labeled documents that were originally in the target language, to learn a target-language classifier.
Some works assume the availability of unlabeled target language documents, and use the tool to translate them into the source language \cite{Wan:2009:CCS:1687878.1687913}, thus transforming the problem into a semi-supervised learning task.

One important problem regarding the black-box translation approach is the closed nature of the main component of the solution. For example, the way that Google Translate operates is not public. We do not know what data resources and what computational resources were used for the translation. Thus, it is very difficult to evaluate and compare performance with such works.

An alternative approach is to map all documents into some Language-Independent Representation (LIR). All works that follow this approach must supply at least three components: a set of language-independent features, a mapping function from the source language to this space, and a similar mapping function from the target language. 

The language-independent features are usually some kind of semantic concepts.  
Some works use Wikipedia concepts \cite{sogaard2015inverted,song2016cross,Ni:2011:CLT:1935826.1935887,franco2014knowledge}, others a set of given topics \cite{Gliozzo:2006:ECC:1220175.1220245}, or a set of matching informative features (one from each domain) \cite{fernandez2016distributional}.
\citeauthor{prettenhofer2011cross} (\citeyear{prettenhofer2011cross}) created a set of language-independent features by first identifying a relatively small set of informative source words in the training documents, and asking an oracle (Google Translate) for the matching words in the target language. 
\citeauthor{fernandez2016distributional} (\citeyear{fernandez2016distributional}) and \citeauthor{zhou2016transfer} (\citeyear{zhou2016transfer}) used similar approaches.

The mapping from the source and target languages to the intermediate representation is usually learned from some external source. For Wikipedia concepts, usually the source is the text of the articles \cite{sogaard2015inverted,song2016cross,Ni:2011:CLT:1935826.1935887,franco2014knowledge}.
Topic-based features and corresponding features are typically inferred from 
comparable corpora \cite{Gliozzo:2006:ECC:1220175.1220245}.
Mapping from each language to these language-independent features is
commonly learned either from the original labeled training set 
\cite{zhou2016transfer,xiao2015semi,li2014learning},
or from independent (unlabeled) corpora in the required language
\cite{prettenhofer2011cross,fernandez2016distributional}.

The work introduced in this paper belongs to a family of works that use language-independent features. Our language-independent features are Wikipedia concepts associated with Wikipedia articles or with Wikipedia categories. We also learn the mapping functions from a comparable corpus (Wikipedia itself) using a method based on Explicit Semantic Analysis \cite{Gabrilovich:2009:WBS}. Unlike most of the works above, we do not apply SVD and retain the original explicit concepts.
Also note that we do not need labeled (or unlabeled) documents from the target language and we can mix any combination of languages in the training and testing sets. Lastly, we overcome the issue of missing documents in the comparable corpus by generating virtual documents based on the ontology hierarchy.

What is common to all the above approaches is that they convert  training or testing documents either into one of the languages or into a language-independent representation. An alternative approach is to learn a classifier based on the training documents in the source language, and convert the classifier into one that can process unlabeled documents in the target language. 

\citeauthor{zhou2014heterogeneous} (\citeyear{zhou2014heterogeneous}), for example, represented each BOW feature from the source language by a linear combination of features from the target language, and used this mapping to transform a well-trained classifier that uses source language features into one that uses target language features.
\citeauthor{xu2017cross} (\citeyear{xu2017cross}) used a parallel corpus to map the classification model from the source domain into the target domain. They used a well-trained source-language classifier to label documents from the parallel corpus, thus obtaining a labeled training set in the target language. Then, they trained a target language classifier on this labeled set. 

\section{Conclusions}
\label{Section:Conclusions}


In this paper, we presented a general framework for generating language-independent features that allow any mixture of languages in the training and testing documents. Our language-independent features are concepts in a given ontology. 
We assume that each concept has associated text in each of the involved languages, and build a language-specific interpreter for each language.  
During training, we use these interpreters to generate language-independent features for each training document. We then train a classifier that is based on these language-independent features.
Given a testing document, we generate features for it and apply the classifier on the language-independent representation.

Two unique components of our solution are related to the hierarchical structure of Wikipedia categories. First, we generate, in addition to the conceptual features, meta-features that are associated with Wikipedia categories. This allows the learning algorithm a further level of abstraction. Second, we exploit the hierarchy to generate virtual articles for languages that do not have them.
We applied our methodology to several multiple-language learning tasks and showed its superiority over the baselines.


One strength of the approach presented here is the fast rate at which our knowledge source, Wikipedia, is expanding. Wikipedia, in all languages, is constantly updated and expanded.  
New events are often reflected in Wikipedia almost instantaneously, which makes Wikipedia a very dynamic and reactive knowledge source.

\clearpage
\bibliography{general}

\begin{thebibliography}{}

\bibitem[\protect\citeauthoryear{Amini and Goutte}{2010}]{amini2010co}
Amini, M.-R., and Goutte, C.
\newblock 2010.
\newblock A co-classification approach to learning from multilingual corpora.
\newblock {\em Machine learning} 79(1-2):105--121.

\bibitem[\protect\citeauthoryear{Amini, Usunier, and Goutte}{2009}]{amaniAUG09}
Amini, M.-R.; Usunier, N.; and Goutte, C.
\newblock 2009.
\newblock Learning from multiple partially observed views - an application to
  multilingual text categorization.
\newblock In {\em Advances in Neural Information Processing Systems 22 (NIPS
  2009)},  28--36.

\bibitem[\protect\citeauthoryear{Bel, Koster, and
  Villegas}{2003}]{Bel03cross-lingualtext}
Bel, N.; Koster, C. H.~A.; and Villegas, M.
\newblock 2003.
\newblock Cross-lingual text categorization.
\newblock In {\em Proc. ECDL 2003},  126--139.

\bibitem[\protect\citeauthoryear{Blei, Ng, and
  Jordan}{2003}]{Blei:2003:LDA:944919.944937}
Blei, D.~M.; Ng, A.~Y.; and Jordan, M.~I.
\newblock 2003.
\newblock Latent dirichlet allocation.
\newblock {\em J. Mach. Learn. Res.} 3:993--1022.

\bibitem[\protect\citeauthoryear{Fern{\'a}ndez, Esuli, and
  Sebastiani}{2016}]{fernandez2016distributional}
Fern{\'a}ndez, A.~M.; Esuli, A.; and Sebastiani, F.
\newblock 2016.
\newblock Distributional correspondence indexing for cross-lingual and
  cross-domain sentiment classification.
\newblock {\em J. Artif. Intell. Res.(JAIR)} 55:131--163.

\bibitem[\protect\citeauthoryear{Franco-Salvador, Rosso, and
  Navigli}{2014}]{franco2014knowledge}
Franco-Salvador, M.; Rosso, P.; and Navigli, R.
\newblock 2014.
\newblock A knowledge-based representation for cross-language document
  retrieval and categorization.
\newblock In {\em EACL}, volume~14,  414--423.

\bibitem[\protect\citeauthoryear{Gabrilovich and
  Markovitch}{2009}]{Gabrilovich:2009:WBS}
Gabrilovich, E., and Markovitch, S.
\newblock 2009.
\newblock Wikipedia-based semantic interpretation for natural language
  processing.
\newblock {\em Journal of Artificial Intelligence Research} 34:443--498.

\bibitem[\protect\citeauthoryear{Gliozzo and
  Strapparava}{2006}]{Gliozzo:2006:ECC:1220175.1220245}
Gliozzo, A., and Strapparava, C.
\newblock 2006.
\newblock Exploiting comparable corpora and bilingual dictionaries for
  cross-language text categorization.
\newblock In {\em Proceedings of the 21st International Conference on
  Computational Linguistics and the 44th Annual Meeting of the Association for
  Computational Linguistics}, ACL-44,  553--560.
\newblock Stroudsburg, PA, USA: Association for Computational Linguistics.

\bibitem[\protect\citeauthoryear{Li \bgroup et al\mbox.\egroup
  }{2014}]{li2014learning}
Li, W.; Duan, L.; Xu, D.; and Tsang, I.~W.
\newblock 2014.
\newblock Learning with augmented features for supervised and semi-supervised
  heterogeneous domain adaptation.
\newblock {\em IEEE transactions on pattern analysis and machine intelligence}
  36(6):1134--1148.

\bibitem[\protect\citeauthoryear{Ling \bgroup et al\mbox.\egroup
  }{2008}]{Ling08canchinese}
Ling, X.; rong Xue, G.; Dai, W.; Jiang, Y.; Yang, Q.; and Yu, Y.
\newblock 2008.
\newblock Can chinese web pages be classified with english data source.
\newblock In {\em In Proceeding of the 17th international conference on World
  Wide Web},  969--978.

\bibitem[\protect\citeauthoryear{Mogadala and
  Rettinger}{2016}]{mogadala2016bilingual}
Mogadala, A., and Rettinger, A.
\newblock 2016.
\newblock Bilingual word embeddings from parallel and non-parallel corpora for
  cross-language text classification.
\newblock In {\em HLT-NAACL},  692--702.

\bibitem[\protect\citeauthoryear{Montalvo \bgroup et al\mbox.\egroup
  }{2007}]{montalvo2007multilingual}
Montalvo, S.; Mart{\'\i}nez, R.; Casillas, A.; and Fresno, V.
\newblock 2007.
\newblock Multilingual news clustering: Feature translation vs. identification
  of cognate named entities.
\newblock {\em Pattern Recognition Letters} 28(16):2305--2311.

\bibitem[\protect\citeauthoryear{Ni \bgroup et al\mbox.\egroup
  }{2011}]{Ni:2011:CLT:1935826.1935887}
Ni, X.; Sun, J.-T.; Hu, J.; and Chen, Z.
\newblock 2011.
\newblock Cross lingual text classification by mining multilingual topics from
  wikipedia.
\newblock In {\em Proceedings of the Fourth ACM International Conference on Web
  Search and Data Mining}, WSDM '11,  375--384.
\newblock New York, NY, USA: ACM.

\bibitem[\protect\citeauthoryear{Olsson, Oard, and
  Haji{\v{c}}}{2005}]{olsson2005cross}
Olsson, J.~S.; Oard, D.~W.; and Haji{\v{c}}, J.
\newblock 2005.
\newblock Cross-language text classification.
\newblock In {\em Proceedings of the 28th annual international ACM SIGIR
  conference on Research and development in information retrieval},  645--646.
\newblock ACM.

\bibitem[\protect\citeauthoryear{Prettenhofer and
  Stein}{2011}]{prettenhofer2011cross}
Prettenhofer, P., and Stein, B.
\newblock 2011.
\newblock Cross-lingual adaptation using structural correspondence learning.
\newblock {\em ACM Transactions on Intelligent Systems and Technology (TIST)}
  3(1):13.

\bibitem[\protect\citeauthoryear{Shi, Mihalcea, and Tian}{2010}]{shi2010cross}
Shi, L.; Mihalcea, R.; and Tian, M.
\newblock 2010.
\newblock Cross language text classification by model translation and
  semi-supervised learning.
\newblock In {\em Proceedings of the 2010 Conference on Empirical Methods in
  Natural Language Processing},  1057--1067.
\newblock Association for Computational Linguistics.

\bibitem[\protect\citeauthoryear{S{\o}gaard \bgroup et al\mbox.\egroup
  }{2015}]{sogaard2015inverted}
S{\o}gaard, A.; Agi{\'c}, {\v{Z}}.; Alonso, H.~M.; Plank, B.; Bohnet, B.; and
  Johannsen, A.
\newblock 2015.
\newblock Inverted indexing for cross-lingual nlp.
\newblock In {\em The 53rd Annual Meeting of the Association for Computational
  Linguistics and the 7th International Joint Conference of the Asian
  Federation of Natural Language Processing (ACL-IJCNLP 2015)},  1713 -- 1722.

\bibitem[\protect\citeauthoryear{Song \bgroup et al\mbox.\egroup
  }{2016}]{song2016cross}
Song, Y.; Upadhyay, S.; Peng, H.; and Roth, D.
\newblock 2016.
\newblock Cross-lingual dataless classification for many languages.
\newblock In {\em IJCAI},  2901--2907.

\bibitem[\protect\citeauthoryear{Steinberger, Pouliquen, and
  Ignat}{2006}]{steinberger2006exploiting}
Steinberger, R.; Pouliquen, B.; and Ignat, C.
\newblock 2006.
\newblock Exploiting multilingual nomenclatures and language-independent text
  features as an interlingua for cross-lingual text analysis applications.
\newblock {\em arXiv preprint cs/0609064}.

\bibitem[\protect\citeauthoryear{Wan}{2009}]{Wan:2009:CCS:1687878.1687913}
Wan, X.
\newblock 2009.
\newblock Co-training for cross-lingual sentiment classification.
\newblock In {\em Proceedings of the Joint Conference of the 47th Annual
  Meeting of the ACL and the 4th International Joint Conference on Natural
  Language Processing of the AFNLP: Volume 1 - Volume 1}, ACL '09,  235--243.
\newblock Stroudsburg, PA, USA: Association for Computational Linguistics.

\bibitem[\protect\citeauthoryear{Xiao and Guo}{2015}]{xiao2015semi}
Xiao, M., and Guo, Y.
\newblock 2015.
\newblock Semi-supervised subspace co-projection for multi-class heterogeneous
  domain adaptation.
\newblock  525--540.

\bibitem[\protect\citeauthoryear{Xu and Yang}{2017}]{xu2017cross}
Xu, R., and Yang, Y.
\newblock 2017.
\newblock Cross-lingual distillation for text classification.
\newblock {\em arXiv preprint arXiv:1705.02073}.

\bibitem[\protect\citeauthoryear{Xu \bgroup et al\mbox.\egroup
  }{2016}]{xu2016cross}
Xu, R.; Yang, Y.; Liu, H.; and Hsi, A.
\newblock 2016.
\newblock Cross-lingual text classification via model translation with limited
  dictionaries.
\newblock In {\em Proceedings of the 25th ACM International on Conference on
  Information and Knowledge Management},  95--104.
\newblock ACM.

\bibitem[\protect\citeauthoryear{Zhou \bgroup et al\mbox.\egroup
  }{2014}]{zhou2014heterogeneous}
Zhou, J.~T.; Tsang, I.~W.; Pan, S.~J.; and Tan, M.
\newblock 2014.
\newblock Heterogeneous domain adaptation for multiple classes.
\newblock  1095--1103.

\bibitem[\protect\citeauthoryear{Zhou \bgroup et al\mbox.\egroup
  }{2016}]{zhou2016transfer}
Zhou, J.~T.; Pan, S.~J.; Tsang, I.~W.; and Ho, S.-S.
\newblock 2016.
\newblock Transfer learning for cross-language text categorization through
  active correspondences construction.
\newblock In {\em AAAI},  2400--2406.

\end{thebibliography}
\bibliographystyle{aaai}
\end{document}